\documentclass[sigconf]{acmart}
\usepackage{multirow}
\AtBeginDocument{%
  }

\setcopyright{acmcopyright}
\copyrightyear{2023}
\acmYear{2023}
\acmDOI{XXXXXXX.XXXXXXX}

\acmConference[ ]{ }{ }{ }

\begin{document}

\title{Interpretable multimodal sentiment analysis based on textual modality descriptions by using large-scale language models}

\author{Sixia Li}
\email{lisixia@jaist.ac.jp}
\affiliation{%
  \institution{Japan Advanced Institute of Science and Technology}
  \city{Nomi}
  \state{Ishikawa}
  \country{Japan}
}

\author{Shogo Okada}
\email{okada-s@jaist.ac.jp}
\affiliation{%
  \institution{Japan Advanced Institute of Science and Technology}
  \city{Nomi}
  \state{Ishikawa}
  \country{Japan}
}

\renewcommand{\shortauthors}{ et al.}

\begin{abstract}
Multimodal sentiment analysis is an important area for understanding the user's internal states. Deep learning methods were effective, but the problem of poor interpretability has gradually gained attention. Previous works have attempted to use attention weights or vector distributions to provide interpretability. However, their explanations were not intuitive and can be influenced by different trained models. This study proposed a novel approach to provide interpretability by converting nonverbal modalities into text descriptions and by using large-scale language models for sentiment predictions. This provides an intuitive approach to directly interpret what models depend on with respect to making decisions from input texts, thus significantly improving interpretability. Specifically, we convert descriptions based on two feature patterns for the audio modality and discrete action units for the facial modality. Experimental results on two sentiment analysis tasks demonstrated that the proposed approach maintained, or even improved effectiveness for sentiment analysis compared to baselines using conventional features, with the highest improvement of 2.49\% on the F1 score. The results also showed that multimodal descriptions have similar characteristics on fusing modalities as those of conventional fusion methods. The results demonstrated that the proposed approach is interpretable and effective for multimodal sentiment analysis.
\end{abstract}




\maketitle

\section{Introduction}

{\let\thefootnote\relax\footnote{The code and experiment data of this study have been open-sourced at: https://github.com/Xia-code/Textual-modality-description}}

Multimodal sentiment analysis in human-computer interaction (HCI) is an important area to understand the user's sentiment state based on multimodal signals. With the development of neural networks, neural network-based methods have been shown high accuracies for multimodal sentiment analysis in obtaining modality representations and in performing modality fusions \cite{soleymani2017survey, han2021bi}.

However, the problem of poor interpretability of how different modalities work has gradually gained attention \cite{joshi2021review}. Due to the black-box characteristics, it is difficult to explain the mechanism of neural networks directly. Previous studies mainly focused on the correspondence between the output of the model and the multimodal information to interpret models, such as analyzing the distribution of output vectors \cite{wang2021m2lens, yuan2021survey} and the weights in attention mechanisms \cite{chen2017recurrent, letarte2018importance, alipour2020study}. However, these methods do not intuitively provide an understanding of the roles of modalities. Moreover, these methods are sensitive to the training of the neural networks. Differently trained models can lead to biased results of the analysis.

Inspired by the generalizability of large-scale language models (LLMs) such as ChatGPT \cite{leiter2023chatgpt} and BERT \cite{kenton2019BERT, lin2021fast} that can understand human descriptions well and the fact that people can estimate others internal states based on language descriptions of behavior and states, we proposed a novel approach for interpretable multimodal sentiment analysis. That is to convert nonverbal modalities into textual descriptions and use LLMs for predicting sentiment based on these descriptions. Figure 1 shows an example of our approach compared to conventional methods. In the example, unlike conventional approaches, our approach converts nonverbal modalities into textual descriptions such as 'the pitch raises then falls' for the audio modality and 'raises eyebrows' for the facial modality, and then our approach utilizes LLMs for sentiment predictions. Specifically, textual modality descriptions are not obtained from image caption \cite{li2022mplug} or attribute learning \cite{ yang2018commonsense}, since captions generally are not related to sentiments. On the other hand, we propose to use descriptions of feature patterns that are related to sentiments, such as how pitch changes for audio modality. The proposed approach provides a paradigm to directly interpret to humans what the model depends on in decision-making from input texts. The proposed approach can also provide a new paradigm to conveniently explore effective modality patterns by adjusting modality descriptions. To the best of our knowledge, we are the first to propose this approach.

\begin{figure}[t]
\centering
\includegraphics[width=7.5cm]{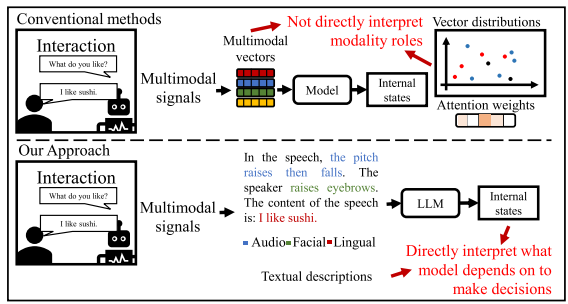}
\caption{Comparison between our approach and previous works}
\end{figure}

To comprehensively verify the effectiveness of the proposed approach, we conducted experiments on two sentiment analysis tasks from two perspectives. (1) One perspective is verifying whether the proposed approach is effective for sentiment analysis while providing interpretabilities. Specifically, we compare models using textual modality descriptions and conventional modality features. (2) The other perspective is exploring whether generally multimodal descriptions generally provide more helpful information than that of single modality descriptions in estimating sentiment. Specifically, we use textual modality descriptions as the prompt and ask ChatGPT (API) for sentiment prediction responses. The results in Section 5.1 to Section 5.3 demonstrated that the proposed approach maintains and even is more effective than that of using conventional features. The results in Section 5.4 showed that multimodal descriptions are generally more helpful information than those provided by single modal descriptions in sentiment predictions based on ChatGPT (API). These results demonstrated that the proposed approach is much more interpretable than that of previous works and is effective for multimodal sentiment analysis.

Our contributions can be summarized as follows:

1. We proposed a novel approach for interpretable multimodal sentiment analysis to convert modality information into textual descriptions and to use LLMs for sentiment predictions. This provides a new paradigm to make model decisions to be directly interpreted by the input texts. (Section 3)

2. We conducted experiments on two sentiment analysis tasks. The results showed that the proposed approach maintained, or even improved effectiveness compared to those of baseline models that using conventional features. (Section 5.1)

3. We used ChatGPT (API) to evaluate our model from another perspective, and the results showed that multimodal descriptions generally provide more helpful information than single modalities in sentiment analysis. (Section 5.2)

4. We compared the performances of two description combination forms and showed that combining modality descriptions in a way that is close to natural texts is better. (Section 5.1)

\section{Related works}
\subsection{Multimodal sentiment analysis}
Multimodal sentiment analysis is an important area of HCI: it aims to extract to analyze public moods and views \cite{gandhi2022multimodal}. Polarity and intensity are two widely used annotations to describe sentiment states in previous studies. Polarity is usually annotated by positive, negative, and natural categories, such as in the MELD dataset \cite{poria2019meld} and the CH-SIMS dataset \cite{yu2020ch}. Intensity is usually annotated in points, such as the 7-point scale in the MOSI dataset \cite{zadeh2016mosi} and the Hazumi dataset \cite{komatani2021multimodal}. Recently, a potential problem with annotation has attracted attention. That is conventional annotations were mainly from the viewpoints of third-parties, but sentiment is an interval state and annotators need to estimate the subject's sentiment from explicit information. This can lead to bias. Therefore, annotations by self-reported sentiment were proposed to describe sentiment closer to the real internal sentiments \cite{katada2020she, hirano2021recognizing}. In this study, we will conduct experiments on both third-party sentiment and self-reported sentiment to evaluate our approach. In multimodal sentiment analysis, text, audio, and video are basic widely used modalities\cite{yang2020cm, mai2022hybrid, chan2023state}. Recently, physiological and physical features were also shown to be effective, such as electroencephalogram \cite{miranda2018amigos} and the electrodermal activity phasic component \cite{katada2022effects}. In representing modality information, early works mainly used hand-crafted features, such as utterance length \cite{okada2016estimating} for the lingual modality, and feature sets \cite{schuller2009interspeech, eyben2015geneva} for the audio modality. Recently, obtaining representations from pretrained model encoding has been shown to be effective \cite{zou2022utilizing, wei2021multimodal}. In fusing modalities, early fusion and late fusion \cite{gadzicki2020early} are popular ways. Recently, complicated fusion methods such as tensor fusion \cite{zadeh2017tensor} and bimodal fusion \cite{han2021bi} were shown to be effective for capturing interaction characteristics among modalities. However, with the vectorization of modality representations and expansion of complicated fusion models, the interpretability of multimodal sentiment analysis has gradually gained attention.

\subsection{Interpretability studies}
Previous works mainly focused on explaining models based on the correspondence between input modalities and the output results. Analyzing attention weights \cite{chen2017recurrent, letarte2018importance, alipour2020study} and visualizing modality vectors have been popular approaches \cite{joshi2021review, wang2021m2lens}. However, these explaining are from the output, they were not intuitively related to inputs and can be influenced by differently trained models.

For these problems, we proposed to convert nonverbal modalities into descriptions that can be directly understandable for humans. In previous studies, relevant efforts converted images into text captions \cite{yang2022few} and correlated audio and text descriptions for emotion recognition \cite{dhamyal2022describing}. However, these studies did not focus on interpretability, and they concatenated other dense representations. On the other hand, we use construct interpretability from the input side, so the model output can be explained regarding to the input textual descriptions alone. Another relevant direction is attribute learning for explaining image classification [33, 34, 35]. Those studies used attributes or sentences to explain which components in the input images were important. However, attributes mainly describe images and not sentiments. On the other hand, our approach converts behaviors into descriptions related to internal states. Thus, our approach can also be applied for counseling psychiatry and assessment of mental disease areas, where descriptions of behaviors related to internal states are important \cite{pampouchidou2017automatic}.

\begin{figure*}[t]
\centering
\includegraphics[width=17cm]{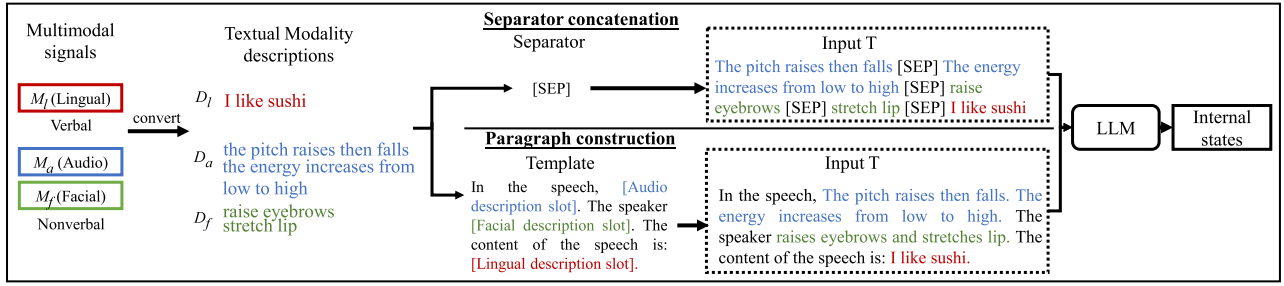}
\caption{The process of converting modality to textual descriptions and description combination methods}
\end{figure*}

\section{Proposed approach}
\subsection{Methodology}
We proposed an interpretable approach to convert nonverbal modalities into textual descriptions and use LLMs to model multimodal sentiment analysis. Figure 2 shows the process of our approach. We formulate the process as follows:

For a given utterance consisted of multiple modalities $U$ = [$M_l$, $M_a$, $M_f$, $...$]. $M_l$, $M_a$, and $M_f$ indicate the raw signal of lingual, audio, and facial modalities, respectively. We convert the nonverbal modalities $M_a$, $M_f$, $...$ into textual descriptions of $D_a$, $D_f$, $...$, respectively. These descriptions are obtained based on modality features or feature patterns. We do not convert the lingual (verbal) modality since it is already in the text form, and we use the $M_l$ as the description of the lingual modality, $D_l$. Then we combine modality descriptions to construct an input $T$ = $D_l$ | $D_a$ | $D_f$ | $...$. The '$|$' indicates the combination method. We use two combination methods for constructing T in this study: they are described in Section 3.3. Then, we use LLMs with the input T to obtain sentiment predictions. For our purpose, we use discriminative LLMs with classification heads and generative LLMs. Details about using LLMs are introduced in Section 3.3.

In this way, we constructed the input T to directly interpret what the model depends on in making decisions. We use the understanding capability of LLMs to predict sentiment based on T.

\subsection{Textual modality descriptions}
We use nonverbal modalities including audio and facial modalities, as they are general modalities that are available in many scenarios. Specifically, we convert the audio modality based on pitch and energy feature patterns, and we convert the facial modality based on action unit features. Choosing these features is because they are basic features that are related to sentiment states and have been shown to be effective in many tasks \cite{katada2020she, wei2021multimodal, wei2022investigating}. Although only focusing on limited features leads to a loss of information of modalities, the purpose of this study is to verify whether the proposed approach can be effective while providing interpretabilities. The loss of information can be complemented in the future by exploring methods relative to comprehensively describing modalities. Moreover, if using information can be as effective as using conventional features, our approach can be promising in terms of efficiency.

\textbf{Audio:} We convert descriptions for the audio modality based on change patterns of pitch and energy in the given speech. We first use the pitch function from MATLAB to estimate the pitch for each frame, and we computed the root mean square energy to obtain the energy for each frame. Then we divide the given speech into three average periods based on the duration, and we compute the average pitch / energy in each period. Finally, we summarize the change patterns among periods in five cases:

a. The average pitch / energy holds or decreases from period 1 to period 2, and decreases from period 2 to period 3; or the average feature decreases from period 1 to period 2, and holds or decreases from period 2 to period 3;

b. The average pitch / energy holds or increases from period 1 to period 2, and increases from period 2 to period 3; or the average feature increases from period 1 to period 2, and holds or increases from period 2 to period 3;

c. The average pitch / energy decreases from period 1 to period 2, and increases from period 2 to period 3;

d. The average pitch / energy increases from period 1 to period 2, and decreases from period 2 to period 3;

e. The average pitch / energy is the same in all periods.

The descriptions for these five cases are shown in Table 1. Based on the process described above, we can convert a given speech into two textual modality descriptions for pitch and energy change patterns respectively.

\begin{table}[t]
\centering
\caption{Descriptions for the audio modality}
\begin{tabular}{c|ll}
\cline{1-2}
Feature pattern & \multicolumn{1}{c}{Modality description}  &  \\ \cline{1-2}
a               & pitch / energy decreases from high to low &  \\
b               & pitch / energy increases from low to high &  \\
c               & pitch / energy rises and then falls       &  \\
d               & pitch / energy falls and then rises       &  \\
e               & pitch / energy does not change            &  \\ \cline{1-2}
\end{tabular}
\end{table}

\textbf{Facial:} We convert descriptions of the facial modality based on action units (AUs) obtained from OpenFace \cite{baltrusaitis2018openface, baltruvsaitis2015cross}. We first extract discrete AUs by OpenFace. If an AU appeared over half of frames in one utterance, we treat that AU as 'appeared.' If no AU is 'appeared', we define a behavior called no appeared AU for such utterances. In particular, OpenFace provides 18 discrete AUs. After checking these 18 AUs, we then obtain descriptions of all 'appeared' AUs from AU descriptions. The original AU descriptions from OpenFace were telling names of AUs, such as 'inner brow raiser.' However, our purpose is to describe the participant behaviors related to sentiment. Therefore, we modify original descriptions slightly into an action description, such as 'raise inner brow.' Table 2 shows modified descriptions for each AU. As a result, we can obtain multiple facial descriptions for one given utterance based on the appeared AUs.

\begin{table}[t]
\centering
\caption{Descriptions of action units for the facial modality}
\begin{tabular}{cl|cl}
\hline
\multicolumn{1}{c|}{AU}   & Description      & \multicolumn{1}{c|}{AU}       & Description           \\ \hline
\multicolumn{1}{c|}{AU1}  & raise inner brow & \multicolumn{1}{c|}{AU14}     & dimple                \\
\multicolumn{1}{c|}{AU2}  & raise outer brow & \multicolumn{1}{c|}{AU15}     & depress lip corner    \\
\multicolumn{1}{c|}{AU4}  & lower brow       & \multicolumn{1}{c|}{AU17}     & raise chin            \\
\multicolumn{1}{c|}{AU5}  & raise upper lid  & \multicolumn{1}{c|}{AU20}     & stretch lip           \\
\multicolumn{1}{c|}{AU6}  & raise cheek      & \multicolumn{1}{c|}{AU23}     & tighten lip           \\
\multicolumn{1}{c|}{AU7}  & tighten lid      & \multicolumn{1}{c|}{AU25}     & part lip              \\
\multicolumn{1}{c|}{AU9}  & wrinkle nose     & \multicolumn{1}{c|}{AU26}     & drop jaw              \\
\multicolumn{1}{c|}{AU10} & raise upper lip  & \multicolumn{1}{c|}{AU28}     & suck lip              \\
\multicolumn{1}{c|}{AU12} & pull lip corner  & \multicolumn{1}{c|}{AU45}     & blink                 \\ \hline \hline
\multicolumn{2}{c|}{Additional}            & \multicolumn{2}{l}{Description}                       \\ \hline
\multicolumn{2}{c|}{No appeared AU}           & \multicolumn{2}{l}{have no obvious facial expression} \\ \hline
\end{tabular}
\end{table}

\subsection{Description combination}
\subsubsection{Discriminative LLM-based method}
\hfill\\
To input modality descriptions into discriminative LLMs, we use two methods for combining modality descriptions into input text T, as shown in Figure 2.

\textbf{Separator concatenation:} We use separators to connect descriptions, as separators are widely used in LLM research \cite{kenton2019BERT, liu2019RoBERTa, wang2021entailment, wang2019structBERT}. The separator concatenation in Figure 2 shows an example of this method. Specifically, we use the separator corresponding to the used LLM (e.g., [SEP] for the BERT model). In the process, we first combine descriptions within modalities (i.e., descriptions of pitch and energy, and different AUs) and then combine descriptions from different modalities in the order of audio, facial, and lingual. For experiments that do not use all three modalities, we combine the chosen modalities following the order above.

\textbf{Paragraph construction:} As audio and facial modalities contain multiple descriptions of pitch, energy, and AUs, separator concatenation tends to use too many separators in combination. Based on the consideration that LLMs were trained on data of real texts (e.g., Wikipedia data), using too many separators potentially makes the input text unnatural and influences model performances. Therefore, we consider another way to combine descriptions into paragraphs to make the input text more natural. The paragraph construction in Figure 2 shows an example of this method. We first make a template with slots to be filled, and then we put modality descriptions into corresponding slots. The order of description slots is the same as the modality order in the separator concatenation method, which is audio, facial, and lingual modalities. In paragraph construction, no separators were used on connections within or among modalities. The complete input text can be expected to be more natural and to avoid potential problems of separators.

\subsubsection{ChatGPT-based method}
\hfill\\
As ChatGPT was trained on a wide range of human texts, if ChatGPT can predict sentiment states more correctly by using multimodal descriptions than by using single modality descriptions, it can demonstrate that multimodal textual descriptions provide more helpful information on sentiment predictions. For this purpose, we construct inputs based on the paragraph construction. Figure 3 shows an input example. In particular, we combine modality descriptions by paragraph construction, and we change the paragraph into a prompt form by adding 'Given a description' in front of the paragraph. Then, we add another prompt to give sentiment categories that need the model to answer, which is 'Given sentiment categories of [high, low]'. Finally, we add a question to ask the model for answers: 'Which sentiment category does the given description belong to?'

\begin{figure}[t]
\centering
\includegraphics[width=8cm]{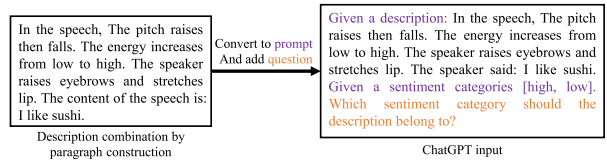}
\caption{The process of constructing inputs for ChatGPT}
\end{figure}

\section{Experiment}
To comprehensively evaluate our approach, we conducted experiments for two multimodal sentiment analysis tasks on the Hazumi dataset \cite{komatani2021multimodal}. One is self-reported sentiment prediction, and the other one is third-party sentiment prediction. Both tasks are predicting user sentiments on the exchange (dialogue turn) level.

\subsection{Dataset}
The Hazumi dataset \cite{komatani2021multimodal} is a publicly available\footnote[1]{https://www.nii.ac.jp/dsc/idr/en/rdata/Hazumi/} Japanese dataset containing many record subsets. This study conducted experiments on the Hazumi1902 and Hazumi1911 subsets. They are only different in recording date and candidates (indicated by 1902 and 1911), and the recoding settings were the same. Data from these two subsets are merged into one set in this study. For convenience, we call the merged dataset as the Hazumi dataset hereafter.

The dataset contains human-computer dialogues from 60 participants. Data include videos from the front side, audio, and lingual transcripts. We extract the transcripts for the lingual modality, pitch and energy for the audio modality, and discrete AUs for the facial modality. As this dataset is in Japanese, all modality descriptions are converted into Japanese for our experiments. For annotations, self-reported and third-party sentiment levels of 1-7 were annotated on each dialogue turn. This study performs experiments on predicting sentiments for these two kinds of labels separately following the settings in previous studies \cite{katada2020she, wei2022investigating}. Specifically, labels 1-7 are compressed into two categories of low and high. Labels 1-4 are treated as low sentiment, and labels 5-7 are treated as high sentiment. Accordingly, this study performs binary classification tasks of predicting self-reported and third-party sentiment. 

Data cleaning was performed by discarding the videos that aligned audio cannot be extracted and the utterances that are missing labels or modality features. After data cleaning, the dataset contains 59 participants with 5091 utterances for our experiments. 

\subsection{Model implementations}
We use discriminative LLMs with a classification head for sentiment predictions based on modality descriptions. We use ChatGPT (API) to generate sentiment predictions based on modality description prompts. Accordingly, experiments using discriminative LLMs are classification tasks in which we can obtain the output directly. Experiments using ChatGPT are generative tasks in which we need to extract the predicted sentiment class from responses. Next, we introduce the process of our experiments.

\textbf{Discriminative LLM models: } We use BERT and RoBERTa for experiments of conventional LLMs since they were shown to be suitable and are widely used on discriminative tasks \cite{letarte2018importance, liu2019RoBERTa, liu2023pre}. As the Hazumi dataset is a Japanese dataset, we use cl-tohoku/BERT-base-japanese for the BERT model and rinna/japanese-RoBERTa-base for the RoBERTa model. These models were pretrained on Japanese. In the model process, we input the description combination and obtain the hidden states of the last layer. We use two pooling methods before classification layers. One method uses the representation of the CLS token since CLS is considered to be the representation of the whole given utterance \cite{kenton2019BERT}. The other method is to compute the mean of token representations among the given utterance, which was shown to be effective in previous works \cite{katada2020she}. Then, we convey the representation after pooling to classification layers to obtain sentiment predictions. To precisely evaluate the effectiveness of textual modality descriptions, we use simple structures to set classification layers as a single immediate layer with 768 units.

\textbf{ChatGPT:} We use the ChatGPT API that corresponds to GPT 3.5 Turbo for experiments. We input prompts with questions to ChatGPT. Then, ChatGPT generates an utterance of answers. In some cases, ChatGPT answers the exact category. But in other cases, it answers that the description does not belong to any of the categories, or the information is not enough to judge which category is correct. Therefore, we perform two strategies to obtain sentiment predictions from ChatGPT's responses. In the case that the answer contains an exact category, we extract the category from the answer as the prediction. In the case that the model did not give a clear answer, we treat the prediction for the corresponding input as an incorrect prediction. In particular, we randomly choose an incorrect label as the prediction for such cases.

\subsection{Baselines}
As we aim to verify the effectiveness of textual modality descriptions compared that of the conventional features, this study uses a single dialogue turn without considering time series. To make fair comparisons, we evaluate our approach by comparing it to four baseline models that also model a single turn. One baseline model is a DNN model that has the same structure to the classification layers at the top of our discriminative LLMs, which contains one immediate layer with 768 units. This model uses early fusion. We use this baseline to make a fair comparison by controlling model structures, so that we can precisely evaluate the effectiveness of textual modality descriptions compared to conventional features. We call this baseline DNN-base hereafter.

The other three baselines are from previous works that used the Hazumi dataset \cite{katada2020she, hirano2021recognizing, hirano2019multitask}, including an early fusion model and two late fusion models. We compare with these baselines to evaluate whether our approach generally performs well or not. The early fusion model contains four immediate layers for encoding fused modalities, and we call this baseline Early Fusion hereafter. In the late fusion baselines, one uses independent immediate layers to encode each modality feature. Then, this approach concatenates encodings together and uses shared immediate layers with a classification layer to obtain predictions. We call this baseline Late Fusion 1 hereafter. The other late fusion model first uses independent immediate layers to encode each modality feature. Then, this second model ensembles the outputs of each modality by summing the outputs together. Finally, a classification layer is used for prediction. We call this baseline Late Fusion 2 hereafter.

For modality features, we followed previous works \cite{katada2020she, wei2021multimodal} and use BERT encoding with mean pooling for the linguistic modality, the InterSpeech 2009 feature set \cite{schuller2009interspeech} for the audio modality, and the mean of discrete AUs for the facial modality. We standardized audio and facial features by using z-scores to convert them into the same scope with the BERT encoding. In the fusion process for baselines, we concatenate modalities in the order of audio, facial, and linguistic when the modality is used. This order is the same as the modality combination for textual modality descriptions.

\subsection{Experimental setting}
\textbf{Training strategy: } We performed 5-fold participant-independent training on the dataset to comprehensively evaluate our approach. In particular, we divide the dataset into five folds on average based on participants. Then, we use one fold as the test set and the other four folds as the training set. In the training set, we split the first 80\% of data for training the model, and we use the last 20\% of data as the validation set to choose the best parameter based on performances. By setting each fold as the test set, we train five separate models for the dataset. The averaged performance on all test sets is used as the final performance for one model setting.
  
For discriminative LLMs, in addition to using different pooling methods, we apply the training settings of freezing and fine-tuning. Specifically, we update the parameters only of the classification layers in the freezing strategy; in the fine-tuning strategy, we update parameters of the pretrained model and the classification layers. We expect to comprehensively evaluate our approach by using multiple training strategies and model settings.

We run all experiments three times to reduce the influence of random initialization. The average performance of three runs is used for evaluation.

\textbf{Parameter setting: }
As described above, we set the classification layers over BERT and RoBERTa model as one immediate layer with 768 units. For baselines, DNN-base was set to be one immediate layer with 768 units. Early Fusion was set to be four immediate layers with 128 units of the first two layers and 64 for the last two layers. Late Fusion 1 was set to be a two-immediate-layer network with 128 units for encoding each modality, and two immediate layers with 64 units for encoding the concatenation of modality encodings. Late Fusion 2 was set to be a separate four-immediate-layer network for encoding each modality, with 128 units of the first two layers and 64 for the last two layers.

In the training process, we performed early stop if the validation metric was not increasing within 5 epochs. We use the cross-entropy loss as the loss function. For fine-tuning, we set the max epoch as 40, the batch size as 16, and the learning rate as 2e-5 throughout the whole training process. For freezing and baseline models, we set the max epoch as 200, the batch size as 32, and we set the learning rate as 0.001 for the first 50 epochs for warming up. We did not perform early stop in the first 50 epochs. Then we decreased the learning rate to 0.0005 for the remaining epochs. We use the macro F1 score as the evaluation metric.

\section{Results and discussions}
\subsection{Discriminative LLM results}

\begin{table*}[]
\centering
\caption{F1 scores of different methods on (a) self-reported sentiment predictions and (b) third-party sentiment predictions}
\begin{tabular}{cccllccccccc}
\multicolumn{12}{l}{(a) F1 results on self-reported sentiment predictions}                                                                                                                                                                                                                                                                               \\ \hline
\multicolumn{4}{c|}{Methods}                                                                                                                                                            & \multicolumn{1}{l|}{Modality}      & A             & F             & A+F           & L             & L+A           & L+F           & L+A+F         \\ \hline
\multicolumn{4}{c|}{\multirow{4}{*}{Baselines}}                                                                                                                                         & \multicolumn{1}{l|}{DNN-base}      & 56.69\%       & 50.71\%       & {\underline {\textbf{58.37\%}}} & {\underline {58.65\%}} & 57.79\%       & {\underline {57.15\%}} & 58.77\%       \\
\multicolumn{4}{c|}{}                                                                                                                                                                   & \multicolumn{1}{l|}{Early Fusion \cite{hirano2019multitask, katada2022effects}}  & 58.72\%       & 50.20\%       & 58.35\%       & 57.65\%       & 57.34\%       & 55.99\%       & {\underline {\textbf{58.80\%}}} \\
\multicolumn{4}{c|}{}                                                                                                                                                                   & \multicolumn{1}{l|}{Late Fusion 1 \cite{hirano2021recognizing, katada2022effects}} & {\underline {\textbf{59.21\%}}} & {\underline {\textbf{51.97\%}}} & 58.27\%       & 58.41\%       & {\underline {58.62\%}} & 56.58\%       & 57.42\%       \\
\multicolumn{4}{c|}{}                                                                                                                                                                   & \multicolumn{1}{l|}{Late Fusion 2 \cite{katada2022effects}} & 58.93\%       & 51.43\%       & 57.45\%       & 58.20\%       & 57.96\%       & 55.20\%       & 58.53\%       \\ \hline
\multicolumn{1}{c|}{\multirow{16}{*}{Ours}} & \multicolumn{1}{c|}{\multirow{8}{*}{SEP}}  & \multicolumn{1}{c|}{\multirow{4}{*}{CLS}}  & \multicolumn{1}{l|}{\multirow{2}{*}{Freeze}}    & \multicolumn{1}{l|}{BERT}          & 54.98\%       & 45.02\%       & 43.91\%       & 57.80\%       & 58.89\%       & 55.24\%       & 57.32\%       \\
\multicolumn{1}{c|}{}                       & \multicolumn{1}{c|}{}                      & \multicolumn{1}{c|}{}                      & \multicolumn{1}{l|}{}                           & \multicolumn{1}{l|}{RoBERTa}       & 54.43\%       & 46.79\%       & 46.83\%       & 55.33\%       & 58.05\%       & 53.77\%       & 56.15\%       \\ \cline{4-12} 
\multicolumn{1}{c|}{}                       & \multicolumn{1}{c|}{}                      & \multicolumn{1}{c|}{}                      & \multicolumn{1}{l|}{\multirow{2}{*}{Fine-tune}} & \multicolumn{1}{l|}{BERT}          & 53.47\%       & 46.02\%       & 47.57\%       & 58.87\%       & 59.96\%       & 55.78\%       & 56.74\%       \\
\multicolumn{1}{c|}{}                       & \multicolumn{1}{c|}{}                      & \multicolumn{1}{c|}{}                      & \multicolumn{1}{l|}{}                           & \multicolumn{1}{l|}{RoBERTa}       & 49.99\%       & 41.21\%       & 43.58\%       & 57.13\%       & 58.73\%       & 56.96\%       & 55.59\%       \\ \cline{3-12} 
\multicolumn{1}{c|}{}                       & \multicolumn{1}{c|}{}                      & \multicolumn{1}{c|}{\multirow{4}{*}{Mean}} & \multicolumn{1}{l|}{\multirow{2}{*}{Freeze}}    & \multicolumn{1}{l|}{BERT}          & 54.98\%       & 42.45\%       & 46.22\%       & 57.40\%       & 59.53\%       & \textbf{57.85\%}       & 57.41\%       \\
\multicolumn{1}{c|}{}                       & \multicolumn{1}{c|}{}                      & \multicolumn{1}{c|}{}                      & \multicolumn{1}{l|}{}                           & \multicolumn{1}{l|}{RoBERTa}       & 54.63\%       & 45.73\%       & 48.01\%       & 56.47\%       & 58.04\%       & 54.51\%       & 54.45\%       \\ \cline{4-12} 
\multicolumn{1}{c|}{}                       & \multicolumn{1}{c|}{}                      & \multicolumn{1}{c|}{}                      & \multicolumn{1}{l|}{\multirow{2}{*}{Fine-tune}} & \multicolumn{1}{l|}{BERT}          & 52.54\%       & 44.81\%       & 47.11\%       & \textbf{59.66\%}       & 58.02\%       & 57.31\%       & 57.74\%       \\
\multicolumn{1}{c|}{}                       & \multicolumn{1}{c|}{}                      & \multicolumn{1}{c|}{}                      & \multicolumn{1}{l|}{}                           & \multicolumn{1}{l|}{RoBERTa}       & 51.30\%       & 39.77\%       & 44.28\%       & 58.38\%       & 59.29\%       & 56.29\%       & 58.67\%       \\ \cline{2-12} 
\multicolumn{1}{c|}{}                       & \multicolumn{1}{c|}{\multirow{8}{*}{Para}} & \multicolumn{1}{c|}{\multirow{4}{*}{CLS}}  & \multicolumn{1}{l|}{\multirow{2}{*}{Freeze}}    & \multicolumn{1}{l|}{BERT}          & 54.98\%       & 41.54\%       & 47.12\%       & 57.65\%       & 58.12\%       & 55.90\%       & 57.28\%       \\
\multicolumn{1}{c|}{}                       & \multicolumn{1}{c|}{}                      & \multicolumn{1}{c|}{}                      & \multicolumn{1}{l|}{}                           & \multicolumn{1}{l|}{RoBERTa}       & 54.99\%       & 44.14\%       & 44.88\%       & 57.41\%       & 58.26\%       & 55.51\%       & 55.63\%       \\ \cline{4-12} 
\multicolumn{1}{c|}{}                       & \multicolumn{1}{c|}{}                      & \multicolumn{1}{c|}{}                      & \multicolumn{1}{l|}{\multirow{2}{*}{Fine-tune}} & \multicolumn{1}{l|}{BERT}          & 52.46\%       & 45.98\%       & 46.27\%       & 58.63\%       & 58.55\%       & 56.35\%       & 57.32\%       \\
\multicolumn{1}{c|}{}                       & \multicolumn{1}{c|}{}                      & \multicolumn{1}{c|}{}                      & \multicolumn{1}{l|}{}                           & \multicolumn{1}{l|}{RoBERTa}       & 48.26\%       & 39.25\%       & 45.96\%       & 58.19\%       & 59.45\%       & 54.88\%       & 56.96\%       \\ \cline{3-12} 
\multicolumn{1}{c|}{}                       & \multicolumn{1}{c|}{}                      & \multicolumn{1}{c|}{\multirow{4}{*}{Mean}} & \multicolumn{1}{l|}{\multirow{2}{*}{Freeze}}    & \multicolumn{1}{l|}{BERT}          & 54.94\%       & 43.20\%       & 46.33\%       & 59.08\%       & \textbf{59.99\%}       & 55.80\%       & 56.79\%       \\
\multicolumn{1}{c|}{}                       & \multicolumn{1}{c|}{}                      & \multicolumn{1}{c|}{}                      & \multicolumn{1}{l|}{}                           & \multicolumn{1}{l|}{RoBERTa}       & 54.92\%       & 43.91\%       & 45.44\%       & 58.93\%       & 59.80\%       & 56.03\%       & 56.14\%       \\ \cline{4-12} 
\multicolumn{1}{c|}{}                       & \multicolumn{1}{c|}{}                      & \multicolumn{1}{c|}{}                      & \multicolumn{1}{l|}{\multirow{2}{*}{Fine-tune}} & \multicolumn{1}{l|}{BERT}          & 54.52\%       & 46.55\%       & 48.25\%       & 58.85\%       & 59.65\%       & 55.22\%       & 57.69\%       \\
\multicolumn{1}{c|}{}                       & \multicolumn{1}{c|}{}                      & \multicolumn{1}{c|}{}                      & \multicolumn{1}{l|}{}                           & \multicolumn{1}{l|}{RoBERTa}       & 54.82\%       & 38.32\%       & 42.03\%       & 57.82\%       & 59.48\%       & 55.90\%       & 58.48\%       \\ \hline
\multicolumn{12}{l}{(b) F1 results on third-party sentiment predictions}                                                                                                                                                                                                                                                                                 \\ \hline
\multicolumn{4}{c|}{Methods}                                                                                                                                                            & \multicolumn{1}{l|}{Modality}      & A             & F             & A+F           & L             & L+A           & L+F           & L+A+F         \\ \hline
\multicolumn{4}{c|}{\multirow{4}{*}{Baselines}}                                                                                                                                         & \multicolumn{1}{l|}{DNN-base}      & 76.98\%       & 65.79\%       & 76.94\%       & {\underline {84.43\%}} & 83.26\%       & 84.16\%       & 83.29\%       \\
\multicolumn{4}{c|}{}                                                                                                                                                                   & \multicolumn{1}{l|}{Early Fusion \cite{hirano2019multitask, katada2022effects}}  & {\underline {\textbf{77.62\%}}} & 66.16\%       & 77.57\%       & 84.19\%       & 83.52\%       & {\underline {84.55\%}} & 84.13\%       \\
\multicolumn{4}{c|}{}                                                                                                                                                                   & \multicolumn{1}{l|}{Late Fusion 1 \cite{hirano2021recognizing, katada2022effects}} & 76.98\%       & {\underline {\textbf{66.68\%}}} & {\underline {\textbf{77.79\%}}} & 84.15\%       & 84.20\%       & 84.04\%       & 84.11\%       \\
\multicolumn{4}{c|}{}                                                                                                                                                                   & \multicolumn{1}{l|}{Late Fusion 2 \cite{katada2022effects}} & 77.37\%       & 66.54\%       & 77.35\%       & {\underline {84.43\%}} & {\underline {84.32\%}} & 84.18\%       & {\underline {84.20\%}} \\ \hline
\multicolumn{1}{c|}{\multirow{16}{*}{Ours}} & \multicolumn{1}{c|}{\multirow{8}{*}{SEP}}  & \multicolumn{1}{c|}{\multirow{4}{*}{CLS}}  & \multicolumn{1}{l|}{\multirow{2}{*}{Freeze}}    & \multicolumn{1}{l|}{BERT}          & 62.01\%       & 47.22\%       & 57.86\%       & 83.75\%       & 83.43\%       & 82.80\%       & 82.70\%       \\
\multicolumn{1}{c|}{}                       & \multicolumn{1}{c|}{}                      & \multicolumn{1}{c|}{}                      & \multicolumn{1}{l|}{}                           & \multicolumn{1}{l|}{RoBERTa}       & 60.79\%       & 47.92\%       & 58.90\%       & 82.09\%       & 81.44\%       & 78.17\%       & 78.01\%       \\ \cline{4-12} 
\multicolumn{1}{c|}{}                       & \multicolumn{1}{c|}{}                      & \multicolumn{1}{c|}{}                      & \multicolumn{1}{l|}{\multirow{2}{*}{Fine-tune}} & \multicolumn{1}{l|}{BERT}          & 62.25\%       & 38.76\%       & 61.29\%       & 84.90\%       & 84.84\%       & 84.54\%       & 84.95\%       \\
\multicolumn{1}{c|}{}                       & \multicolumn{1}{c|}{}                      & \multicolumn{1}{c|}{}                      & \multicolumn{1}{l|}{}                           & \multicolumn{1}{l|}{RoBERTa}       & 60.04\%       & 34.36\%       & 56.54\%       & 85.27\%       & 85.40\%       & 85.19\%       & 85.48\%       \\ \cline{3-12} 
\multicolumn{1}{c|}{}                       & \multicolumn{1}{c|}{}                      & \multicolumn{1}{c|}{\multirow{4}{*}{Mean}} & \multicolumn{1}{l|}{\multirow{2}{*}{Freeze}}    & \multicolumn{1}{l|}{BERT}          & 61.59\%       & 48.09\%       & 59.26\%       & 84.14\%       & 84.23\%       & 83.80\%       & 83.63\%       \\
\multicolumn{1}{c|}{}                       & \multicolumn{1}{c|}{}                      & \multicolumn{1}{c|}{}                      & \multicolumn{1}{l|}{}                           & \multicolumn{1}{l|}{RoBERTa}       & 60.29\%       & 46.18\%       & 59.43\%       & 82.00\%       & 81.98\%       & 78.94\%       & 79.63\%       \\ \cline{4-12} 
\multicolumn{1}{c|}{}                       & \multicolumn{1}{c|}{}                      & \multicolumn{1}{c|}{}                      & \multicolumn{1}{l|}{\multirow{2}{*}{Fine-tune}} & \multicolumn{1}{l|}{BERT}          & 62.25\%       & 37.54\%       & 61.65\%       & 84.85\%       & 84.47\%       & 84.89\%       & 84.99\%       \\
\multicolumn{1}{c|}{}                       & \multicolumn{1}{c|}{}                      & \multicolumn{1}{c|}{}                      & \multicolumn{1}{l|}{}                           & \multicolumn{1}{l|}{RoBERTa}       & 62.14\%       & 34.36\%       & 56.68\%       & 85.43\%       & 85.48\%       & 85.38\%       & 85.15\%       \\ \cline{2-12} 
\multicolumn{1}{c|}{}                       & \multicolumn{1}{c|}{\multirow{8}{*}{Para}} & \multicolumn{1}{c|}{\multirow{4}{*}{CLS}}  & \multicolumn{1}{l|}{\multirow{2}{*}{Freeze}}    & \multicolumn{1}{l|}{BERT}          & 61.41\%       & 45.92\%       & 58.85\%       & 83.45\%       & 84.04\%       & 83.70\%       & 83.17\%       \\
\multicolumn{1}{c|}{}                       & \multicolumn{1}{c|}{}                      & \multicolumn{1}{c|}{}                      & \multicolumn{1}{l|}{}                           & \multicolumn{1}{l|}{RoBERTa}       & 62.25\%       & 44.89\%       & 60.04\%       & 82.95\%       & 83.20\%       & 82.04\%       & 81.91\%       \\ \cline{4-12} 
\multicolumn{1}{c|}{}                       & \multicolumn{1}{c|}{}                      & \multicolumn{1}{c|}{}                      & \multicolumn{1}{l|}{\multirow{2}{*}{Fine-tune}} & \multicolumn{1}{l|}{BERT}          & 61.13\%       & 40.82\%       & 61.79\%       & 85.12\%       & 85.15\%       & 84.54\%       & 85.00\%       \\
\multicolumn{1}{c|}{}                       & \multicolumn{1}{c|}{}                      & \multicolumn{1}{c|}{}                      & \multicolumn{1}{l|}{}                           & \multicolumn{1}{l|}{RoBERTa}       & 62.25\%       & 35.03\%       & 61.83\%       & \textbf{85.63\%}       & \textbf{85.62\%}       & 85.56\%       & \textbf{85.86\%}       \\ \cline{3-12} 
\multicolumn{1}{c|}{}                       & \multicolumn{1}{c|}{}                      & \multicolumn{1}{c|}{\multirow{4}{*}{Mean}} & \multicolumn{1}{l|}{\multirow{2}{*}{Freeze}}    & \multicolumn{1}{l|}{BERT}          & 61.90\%       & 44.41\%       & 59.54\%       & 84.15\%       & 83.98\%       & 83.90\%       & 84.03\%       \\
\multicolumn{1}{c|}{}                       & \multicolumn{1}{c|}{}                      & \multicolumn{1}{c|}{}                      & \multicolumn{1}{l|}{}                           & \multicolumn{1}{l|}{RoBERTa}       & 61.95\%       & 43.99\%       & 59.68\%       & 83.96\%       & 83.40\%       & 82.15\%       & 82.00\%       \\ \cline{4-12} 
\multicolumn{1}{c|}{}                       & \multicolumn{1}{c|}{}                      & \multicolumn{1}{c|}{}                      & \multicolumn{1}{l|}{\multirow{2}{*}{Fine-tune}} & \multicolumn{1}{l|}{BERT}          & 62.25\%       & 40.97\%       & 61.40\%       & 84.99\%       & 84.55\%       & 85.03\%       & 84.62\%       \\
\multicolumn{1}{c|}{}                       & \multicolumn{1}{c|}{}                      & \multicolumn{1}{c|}{}                      & \multicolumn{1}{l|}{}                           & \multicolumn{1}{l|}{RoBERTa}       & 62.19\%       & 34.36\%       & 60.25\%       & 85.55\%       & 85.25\%       & \textbf{85.71\%}       & 85.35\%       \\ \hline
\end{tabular}
\end{table*}

Table 3 shows the F1 score of different methods. Table 3 (a) shows the results of self-sentiment prediction, and Table 3 (b) shows the results of third-party sentiment prediction. CLS and Mean indicate the models using CLS and mean pooling methods. SEP and Para indicate the modality description combination methods of separator concatenation and paragraph construction. Freeze and fine-tune indicate model training by freezing and fine-tuning LLMs. L, A, and F indicate lingual, audio, and facial modalities. Underlined numbers indicate the best baseline performance, and bold numbers indicate the best performance using each modality.

\textbf{Comparison between our approach and baselines:}
First, we investigate the effectiveness of textual modality descriptions compared to that of the conventional features. By comparing the performances between our models and DNN-base in Table 3 (a), one can see that when using L+A, our models with all settings outperformed DNN-base, with the highest improvement of 2.19\%. Relatively fewer of our models when using L and L+F outperformed DNN-base, with the highest improvements of 1.01\% and 0.70\%, respectively. Our models using L+A+F did not outperform DNN-base, but achieved close performance, with the smallest gap of 0.10\%. By comparing performances between our models with DNN-base of third-party sentiment prediction in Table 3 (b), one can see that when using modalities containing L, more than a half of our models outperformed baselines, with the highest improvements of 1.42\%, 1.86\%, 1.16\%, and 2.49\% on L, L+A, L+F, and L+A+F, respectively. These results demonstrated that textual modality descriptions are effective for multimodal sentiment analysis.

Note that it seems strange that when using the L modality, our models did not perform well compared to DNN-base on self-reported sentiment, while the classification layers of our model is the same as DNN-base. By comparing performances of RoBERTa and BERT, we speculate that the reason is that RoBERTa encodings for the L modality are not as effective as BERT, so RoBERTa models degrade most cases on self-reported sentiment using L alone.

Then, we investigate if our approach generally performs well comparing to the best baselines. As seen in Table 3 (a), when using L+A, over half of our model outperformed the best baseline, with the highest improvement of 1.37\%. As seen in Table 3 (b), when using L, L+A, and L+A+F modalities, over half of our models outperformed the best baseline models, with the highest improvements of 1.20\%, 1.30\%, 1.66\% when using L, L+A, and L+A+F modalities, respectively. These results demonstrated the general effectiveness of textual modality descriptions for multimodal sentiment analysis. These results also suggested that textual modality descriptions have the potential to further improve the performance of sentiment analysis by designing suitable structures, meanwhile these descriptions can somehow provide interpretability based on texts regardless of how modeling structures are complicated.

On the other hand, our model did not outperform baselines when using A, F and A+F for self-reported sentiment and third-party sentiment predictions. For the F and A+F modalities, we speculate that the reason is that the textual descriptions of AUs are not natural texts, as they describe detailed actions of specific facial parts. In daily life, we rarely describe other's facial expressions by saying what parts on the face acted but describing overall feelings of the facial movements. Therefore, detailed descriptions of AUs are not effective for LLMs in modeling sentiment predictions. Describing facial actions more naturally can be considered in future works.

For modality A, the main reason is that we used descriptions of only two audio feature patterns, while the conventional feature set contains comprehensive features for describing audio characteristics. However, notably, our models that use audio modality achieved relatively close performance to baseline models on the self-reported sentiment prediction, with the closest gap of 4.22\% compared to that of the best baseline. As self-reported sentiment tends to be an internal state that is more difficult to estimate from explicit information than third-party sentiment \cite{katada2020she}, the results suggested that by abstracting audio patterns, textual modality descriptions are potentially effective cues for estimating internal states.

\textbf{Comparison between single and multiple modalities: }
Then, we investigate whether multimodal descriptions improve performances compared to that of the single modality descriptions. By comparing performances when using A, F, and A+F modalities, one can see that using A+F modality improved performances when using the F modality, although A+F did not improve performances when using the A modality. By comparing the performances of the L, L+A, L+F, and L+A+F modalities, one can see that most of our models using L+A modalities improved the performances when using the L modality alone on self-reported sentiment. However, relatively few of our models that use L+F and L+A+F improved the performances of using the L modality for self-reported sentiment and third-party sentiment predictions. As discussed above, facial modality descriptions were not natural in communications, and we speculate that the combination of the facial modality degrades the performances of multimodal descriptions (A+F, L+F, and L+A+F) compared to that of the single modality descriptions. On the other hand, audio descriptions were showed better cues for self-reported sentiment predictions. Thus, using audio and facial modalities improved the performances of using facial modality alone, and combing lingual and audio modalities improved the performances of using the lingual modality alone.

By comparing the performances of using each modality among our models and baselines, one can see that performance changes among modalities have similar tendencies between our models and baselines. Therefore, these results suggested that the combination of textual modality descriptions has similar characteristics to the fusion of conventional features: each single modality influences the multimodal effectiveness, and the combination of effective modalities improves single modality performances. Therefore, textual modality descriptions are effective in reflecting each modality's role in modeling and are directly interpretable to humans.

Another notable point is models that use L+A, where baseline models used a feature set and our approach used only two feature pattern descriptions. Our models that use L+A outperformed baseline models in most cases. These results suggested that textual modality descriptions are potentially better fusion methods for specific modality features.

\textbf{Comparison between description combination methods: }
Next, we investigate the effective combination form by comparing the performances of separator concatenation and paragraph construction. From Table 3 (a), one can see that no single combination method leads performances for self-reported sentiment predictions. From Table 3 (b), one can see that most of models with paragraph construction outperformed models with separator concatenation for third-party sentiment predictions. Since self-reported sentiment is more difficult to estimate by others' observations, these results suggested that combination forms matter less than the information involved in predicting self-reported sentiment. On the other hand, the third-party sentiment is annotated based on explicit information, and these results demonstrated that constructing modality descriptions in a way close to natural languages is more effective than simply concatenating descriptions together for predicting third-party sentiment. As multimodal sentiment analysis is generally annotated by third-party observers, the results also suggested that paragraph construction can be more effective than separator concatenation for other tasks in multimodal sentiment analysis.

\subsection{ChatGPT experiment results}

Finally, we investigate whether generally multimodal descriptions can provide more helpful information than single modality descriptions. Table 4 shows the F1 score results of ChatGPT. We do not compare the performance by ChatGPT to other baselines because ChatGPT experiments were in different processes. In Table 4, self-reported and third-party sentiment predictions are indicated.

\begin{table}[t]
\centering
\caption{Results of sentiment predictions by ChatGPT}
\begin{tabular}{l|cccl}
\hline
Modality      & A       & F       & A+F     &                             \\ \hline
Self-reported & 44.30\% & 40.26\% & 45.85\% &                             \\
Third-party   & 41.20\% & 43.15\% & 44.90\% &                             \\ \hline
Modality      & L       & L+A     & L+F     & \multicolumn{1}{c}{L+A+F}   \\ \hline
Self-reported & 35.47\% & 41.48\% & 36.92\% & \multicolumn{1}{c}{43.44\%} \\
Third-party   & 35.15\% & 37.64\% & 36.63\% & \multicolumn{1}{c}{39.81\%} \\ \hline
\end{tabular}
\end{table}

As seen in Table 4, using A+F improves the performances of using single modalities A and F on self-reported sentiment prediction by 1.55\% and 5.60\%, respectively. A+F improved A and F by 3.70\% and 1.75\% on third-party prediction, respectively. L+A, L+F, and L+A+F improve the performances of the single modality L by 6.01\%, 1.46\%, and 7.97\% on self-reported sentiment prediction and improve L by 2.49\%, 1.49\%, and 4.66\% on third-party prediction. We note that using modalities containing L did not perform better than using modalities without L as that in discriminative LLM experiments. We speculate that the lingual modality is less clearly corresponds to sentiment states than the audio and facial modalities. A given utterance can be observed in different sentiment states. Therefore, ChatGPT cannot give answers based on lingual descriptions as clearly as those based on audio and facial descriptions. However, these results showed that multimodal descriptions can provide additional cues to improve single modality performance. Therefore, these results demonstrated that multimodal textual descriptions are generally more effective than single modality descriptions.

\subsection{Case Study}
We show an example to intuitively explain how our approach provides interpretability for sentiment analysis. Figure 4 shows inputs and self-sentiment prediction examples of our approach using textual modality descriptions and the approach using conventional features. The input is the L+A+F modality. As seen in the figure, in the conventional approach, because modalities were converted into vectors and concatenated together, it is difficult to interpret the relationships between prediction and modality components or patterns. On the other hand, the input of our approach is the text that describes the contents or characteristics of modalities. We can directly know that the model predicted the sentiment label of high based on the 'It's real' of the lingual modality, the 'pitch falls then raises; energy decreases from high to low.' of the audio modality, and 'raises cheek,' 'tightens lid,' 'raises upper lip,' and 'pulls lip corner' of the facial modality. This example confirmed the interpretability of the proposed approach.

\begin{figure}[t]
\centering
\includegraphics[width=7.5cm]{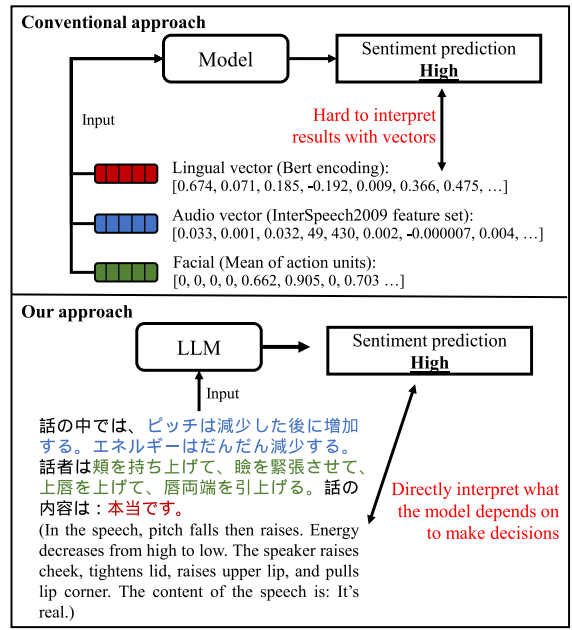}
\caption{Example of interpretability of our approach compared to previous approach}
\end{figure}

\section{Conclusion}
In this study, we proposed a novel approach for interpretable multimodal sentiment analysis by converting nonverbal modalities into textual descriptions and used LLMs for sentiment predictions with textual modality descriptions. The results of discriminative LLMs demonstrated that textual multimodal descriptions maintained, or even improved effectiveness compared to that of the conventional features on two sentiment analysis tasks. Moreover, textual descriptions have similar characteristics in fusing modalities as conventional fusion methods. The results of generative LLMs demonstrated that multimodal descriptions generally provide more helpful information for estimating sentiments than single modality descriptions. The results also suggested that combining modality descriptions in a way that is close to natural communications is better than concatenation by separators. Based on the proposed approach, we can further build models that can explain decisions by themselves by outputting which components are important. This can provide potential applications for diagnostic robots.

On the other hand, our model considered audio and facial nonverbal modalities only, taking other effective modalities, such as head activities and gaze movements, into account can be considered in future works. Moreover, exploring how to comprehensively describe modality information and how to harmonize descriptions with LLMs are also future directions.

\begin{acks}
---
\end{acks}

\bibliographystyle{ACM-Reference-Format}
\bibliography{mybib}


\begin{thebibliography}{43}


\ifx \showCODEN    \undefined \def \showCODEN     #1{\unskip}     \fi
\ifx \showDOI      \undefined \def \showDOI       #1{#1}\fi
\ifx \showISBNx    \undefined \def \showISBNx     #1{\unskip}     \fi
\ifx \showISBNxiii \undefined \def \showISBNxiii  #1{\unskip}     \fi
\ifx \showISSN     \undefined \def \showISSN      #1{\unskip}     \fi
\ifx \showLCCN     \undefined \def \showLCCN      #1{\unskip}     \fi
\ifx \shownote     \undefined \def \shownote      #1{#1}          \fi
\ifx \showarticletitle \undefined \def \showarticletitle #1{#1}   \fi
\ifx \showURL      \undefined \def \showURL       {\relax}        \fi
\providecommand\bibfield[2]{#2}
\providecommand\bibinfo[2]{#2}
\providecommand\natexlab[1]{#1}
\providecommand\showeprint[2][]{arXiv:#2}

\bibitem[Alipour et~al\mbox{.}(2020)]%
        {alipour2020study}
\bibfield{author}{\bibinfo{person}{Kamran Alipour}, \bibinfo{person}{Jurgen~P
  Schulze}, \bibinfo{person}{Yi Yao}, \bibinfo{person}{Avi Ziskind}, {and}
  \bibinfo{person}{Giedrius Burachas}.} \bibinfo{year}{2020}\natexlab{}.
\newblock \showarticletitle{A study on multimodal and interactive explanations
  for visual question answering}.
\newblock \bibinfo{journal}{\emph{arXiv preprint arXiv:2003.00431}}
  (\bibinfo{year}{2020}).
\newblock


\bibitem[Baltru{\v{s}}aitis et~al\mbox{.}(2015)]%
        {baltruvsaitis2015cross}
\bibfield{author}{\bibinfo{person}{Tadas Baltru{\v{s}}aitis},
  \bibinfo{person}{Marwa Mahmoud}, {and} \bibinfo{person}{Peter Robinson}.}
  \bibinfo{year}{2015}\natexlab{}.
\newblock \showarticletitle{Cross-dataset learning and person-specific
  normalisation for automatic action unit detection}. In
  \bibinfo{booktitle}{\emph{2015 11th IEEE International Conference and
  Workshops on Automatic Face and Gesture Recognition (FG)}},
  Vol.~\bibinfo{volume}{6}. IEEE, \bibinfo{pages}{1--6}.
\newblock


\bibitem[Baltrusaitis et~al\mbox{.}(2018)]%
        {baltrusaitis2018openface}
\bibfield{author}{\bibinfo{person}{Tadas Baltrusaitis}, \bibinfo{person}{Amir
  Zadeh}, \bibinfo{person}{Yao~Chong Lim}, {and}
  \bibinfo{person}{Louis-Philippe Morency}.} \bibinfo{year}{2018}\natexlab{}.
\newblock \showarticletitle{Openface 2.0: Facial behavior analysis toolkit}. In
  \bibinfo{booktitle}{\emph{2018 13th IEEE international conference on
  automatic face \& gesture recognition (FG 2018)}}. IEEE,
  \bibinfo{pages}{59--66}.
\newblock


\bibitem[Chan et~al\mbox{.}(2023)]%
        {chan2023state}
\bibfield{author}{\bibinfo{person}{Jireh Yi-Le Chan},
  \bibinfo{person}{Khean~Thye Bea}, \bibinfo{person}{Steven Mun~Hong Leow},
  \bibinfo{person}{Seuk~Wai Phoong}, {and} \bibinfo{person}{Wai~Khuen Cheng}.}
  \bibinfo{year}{2023}\natexlab{}.
\newblock \showarticletitle{State of the art: a review of sentiment analysis
  based on sequential transfer learning}.
\newblock \bibinfo{journal}{\emph{Artificial Intelligence Review}}
  \bibinfo{volume}{56}, \bibinfo{number}{1} (\bibinfo{year}{2023}),
  \bibinfo{pages}{749--780}.
\newblock


\bibitem[Chen et~al\mbox{.}(2017)]%
        {chen2017recurrent}
\bibfield{author}{\bibinfo{person}{Peng Chen}, \bibinfo{person}{Zhongqian Sun},
  \bibinfo{person}{Lidong Bing}, {and} \bibinfo{person}{Wei Yang}.}
  \bibinfo{year}{2017}\natexlab{}.
\newblock \showarticletitle{Recurrent attention network on memory for aspect
  sentiment analysis}. In \bibinfo{booktitle}{\emph{Proceedings of the 2017
  conference on empirical methods in natural language processing}}.
  \bibinfo{pages}{452--461}.
\newblock


\bibitem[Dhamyal et~al\mbox{.}(2022)]%
        {dhamyal2022describing}
\bibfield{author}{\bibinfo{person}{Hira Dhamyal}, \bibinfo{person}{Benjamin
  Elizalde}, \bibinfo{person}{Soham Deshmukh}, \bibinfo{person}{Huaming Wang},
  \bibinfo{person}{Bhiksha Raj}, {and} \bibinfo{person}{Rita Singh}.}
  \bibinfo{year}{2022}\natexlab{}.
\newblock \showarticletitle{Describing emotions with acoustic property prompts
  for speech emotion recognition}.
\newblock \bibinfo{journal}{\emph{arXiv preprint arXiv:2211.07737}}
  (\bibinfo{year}{2022}).
\newblock


\bibitem[Eyben et~al\mbox{.}(2015)]%
        {eyben2015geneva}
\bibfield{author}{\bibinfo{person}{Florian Eyben}, \bibinfo{person}{Klaus~R
  Scherer}, \bibinfo{person}{Bj{\"o}rn~W Schuller}, \bibinfo{person}{Johan
  Sundberg}, \bibinfo{person}{Elisabeth Andr{\'e}}, \bibinfo{person}{Carlos
  Busso}, \bibinfo{person}{Laurence~Y Devillers}, \bibinfo{person}{Julien
  Epps}, \bibinfo{person}{Petri Laukka}, \bibinfo{person}{Shrikanth~S
  Narayanan}, {et~al\mbox{.}}} \bibinfo{year}{2015}\natexlab{}.
\newblock \showarticletitle{The Geneva minimalistic acoustic parameter set
  (GeMAPS) for voice research and affective computing}.
\newblock \bibinfo{journal}{\emph{IEEE transactions on affective computing}}
  \bibinfo{volume}{7}, \bibinfo{number}{2} (\bibinfo{year}{2015}),
  \bibinfo{pages}{190--202}.
\newblock


\bibitem[Gadzicki et~al\mbox{.}(2020)]%
        {gadzicki2020early}
\bibfield{author}{\bibinfo{person}{Konrad Gadzicki}, \bibinfo{person}{Razieh
  Khamsehashari}, {and} \bibinfo{person}{Christoph Zetzsche}.}
  \bibinfo{year}{2020}\natexlab{}.
\newblock \showarticletitle{Early vs late fusion in multimodal convolutional
  neural networks}. In \bibinfo{booktitle}{\emph{2020 IEEE 23rd international
  conference on information fusion (FUSION)}}. IEEE, \bibinfo{pages}{1--6}.
\newblock


\bibitem[Gandhi et~al\mbox{.}(2022)]%
        {gandhi2022multimodal}
\bibfield{author}{\bibinfo{person}{Ankita Gandhi}, \bibinfo{person}{Kinjal
  Adhvaryu}, \bibinfo{person}{Soujanya Poria}, \bibinfo{person}{Erik Cambria},
  {and} \bibinfo{person}{Amir Hussain}.} \bibinfo{year}{2022}\natexlab{}.
\newblock \showarticletitle{Multimodal sentiment analysis: A systematic review
  of history, datasets, multimodal fusion methods, applications, challenges and
  future directions}.
\newblock \bibinfo{journal}{\emph{Information Fusion}} (\bibinfo{year}{2022}).
\newblock


\bibitem[Han et~al\mbox{.}(2021)]%
        {han2021bi}
\bibfield{author}{\bibinfo{person}{Wei Han}, \bibinfo{person}{Hui Chen},
  \bibinfo{person}{Alexander Gelbukh}, \bibinfo{person}{Amir Zadeh},
  \bibinfo{person}{Louis-philippe Morency}, {and} \bibinfo{person}{Soujanya
  Poria}.} \bibinfo{year}{2021}\natexlab{}.
\newblock \showarticletitle{Bi-bimodal modality fusion for
  correlation-controlled multimodal sentiment analysis}. In
  \bibinfo{booktitle}{\emph{Proceedings of the 2021 International Conference on
  Multimodal Interaction}}. \bibinfo{pages}{6--15}.
\newblock


\bibitem[Hirano et~al\mbox{.}(2021)]%
        {hirano2021recognizing}
\bibfield{author}{\bibinfo{person}{Yuki Hirano}, \bibinfo{person}{Shogo Okada},
  {and} \bibinfo{person}{Kazunori Komatani}.} \bibinfo{year}{2021}\natexlab{}.
\newblock \showarticletitle{Recognizing Social Signals with Weakly Supervised
  Multitask Learning for Multimodal Dialogue Systems}. In
  \bibinfo{booktitle}{\emph{Proceedings of the 2021 International Conference on
  Multimodal Interaction}}. \bibinfo{pages}{141--149}.
\newblock


\bibitem[Hirano et~al\mbox{.}(2019)]%
        {hirano2019multitask}
\bibfield{author}{\bibinfo{person}{Yuki Hirano}, \bibinfo{person}{Shogo Okada},
  \bibinfo{person}{Haruto Nishimoto}, {and} \bibinfo{person}{Kazunori
  Komatani}.} \bibinfo{year}{2019}\natexlab{}.
\newblock \showarticletitle{Multitask prediction of exchange-level annotations
  for multimodal dialogue systems}. In \bibinfo{booktitle}{\emph{2019
  International Conference on Multimodal Interaction}}.
  \bibinfo{pages}{85--94}.
\newblock


\bibitem[Joshi et~al\mbox{.}(2021)]%
        {joshi2021review}
\bibfield{author}{\bibinfo{person}{Gargi Joshi}, \bibinfo{person}{Rahee
  Walambe}, {and} \bibinfo{person}{Ketan Kotecha}.}
  \bibinfo{year}{2021}\natexlab{}.
\newblock \showarticletitle{A review on explainability in multimodal deep
  neural nets}.
\newblock \bibinfo{journal}{\emph{IEEE Access}}  \bibinfo{volume}{9}
  (\bibinfo{year}{2021}), \bibinfo{pages}{59800--59821}.
\newblock


\bibitem[Katada et~al\mbox{.}(2020)]%
        {katada2020she}
\bibfield{author}{\bibinfo{person}{Shun Katada}, \bibinfo{person}{Shogo Okada},
  \bibinfo{person}{Yuki Hirano}, {and} \bibinfo{person}{Kazunori Komatani}.}
  \bibinfo{year}{2020}\natexlab{}.
\newblock \showarticletitle{Is she truly enjoying the conversation? analysis of
  physiological signals toward adaptive dialogue systems}. In
  \bibinfo{booktitle}{\emph{Proceedings of the 2020 International Conference on
  Multimodal Interaction}}. \bibinfo{pages}{315--323}.
\newblock


\bibitem[Katada et~al\mbox{.}(2022)]%
        {katada2022effects}
\bibfield{author}{\bibinfo{person}{Shun Katada}, \bibinfo{person}{Shogo Okada},
  {and} \bibinfo{person}{Kazunori Komatani}.} \bibinfo{year}{2022}\natexlab{}.
\newblock \showarticletitle{Effects of Physiological Signals in Different Types
  of Multimodal Sentiment Estimation}.
\newblock \bibinfo{journal}{\emph{IEEE Transactions on Affective Computing}}
  (\bibinfo{year}{2022}).
\newblock


\bibitem[Kenton and Toutanova(2019)]%
        {kenton2019BERT}
\bibfield{author}{\bibinfo{person}{Jacob Devlin Ming-Wei~Chang Kenton} {and}
  \bibinfo{person}{Lee~Kristina Toutanova}.} \bibinfo{year}{2019}\natexlab{}.
\newblock \showarticletitle{BERT: Pre-training of Deep Bidirectional
  Transformers for Language Understanding}. In
  \bibinfo{booktitle}{\emph{Proceedings of NAACL-HLT}}.
  \bibinfo{pages}{4171--4186}.
\newblock


\bibitem[Komatani and Okada(2021)]%
        {komatani2021multimodal}
\bibfield{author}{\bibinfo{person}{Kazunori Komatani} {and}
  \bibinfo{person}{Shogo Okada}.} \bibinfo{year}{2021}\natexlab{}.
\newblock \showarticletitle{Multimodal human-agent dialogue corpus with
  annotations at utterance and dialogue levels}. In
  \bibinfo{booktitle}{\emph{2021 9th International Conference on Affective
  Computing and Intelligent Interaction (ACII)}}. IEEE, \bibinfo{pages}{1--8}.
\newblock


\bibitem[Leiter et~al\mbox{.}(2023)]%
        {leiter2023chatgpt}
\bibfield{author}{\bibinfo{person}{Christoph Leiter}, \bibinfo{person}{Ran
  Zhang}, \bibinfo{person}{Yanran Chen}, \bibinfo{person}{Jonas Belouadi},
  \bibinfo{person}{Daniil Larionov}, \bibinfo{person}{Vivian Fresen}, {and}
  \bibinfo{person}{Steffen Eger}.} \bibinfo{year}{2023}\natexlab{}.
\newblock \showarticletitle{ChatGPT: A Meta-Analysis after 2.5 Months}.
\newblock \bibinfo{journal}{\emph{arXiv preprint arXiv:2302.13795}}
  (\bibinfo{year}{2023}).
\newblock


\bibitem[Letarte et~al\mbox{.}(2018)]%
        {letarte2018importance}
\bibfield{author}{\bibinfo{person}{Ga{\"e}l Letarte},
  \bibinfo{person}{Fr{\'e}d{\'e}rik Paradis}, \bibinfo{person}{Philippe
  Gigu{\`e}re}, {and} \bibinfo{person}{Fran{\c{c}}ois Laviolette}.}
  \bibinfo{year}{2018}\natexlab{}.
\newblock \showarticletitle{Importance of self-attention for sentiment
  analysis}. In \bibinfo{booktitle}{\emph{Proceedings of the 2018 EMNLP
  Workshop BlackboxNLP: Analyzing and Interpreting Neural Networks for NLP}}.
  \bibinfo{pages}{267--275}.
\newblock


\bibitem[Li et~al\mbox{.}(2022)]%
        {li2022mplug}
\bibfield{author}{\bibinfo{person}{Chenliang Li}, \bibinfo{person}{Haiyang Xu},
  \bibinfo{person}{Junfeng Tian}, \bibinfo{person}{Wei Wang},
  \bibinfo{person}{Ming Yan}, \bibinfo{person}{Bin Bi}, \bibinfo{person}{Jiabo
  Ye}, \bibinfo{person}{Hehong Chen}, \bibinfo{person}{Guohai Xu},
  \bibinfo{person}{Zheng Cao}, {et~al\mbox{.}}}
  \bibinfo{year}{2022}\natexlab{}.
\newblock \showarticletitle{mPLUG: Effective and Efficient Vision-Language
  Learning by Cross-modal Skip-connections}.
\newblock \bibinfo{journal}{\emph{arXiv preprint arXiv:2205.12005}}
  (\bibinfo{year}{2022}).
\newblock


\bibitem[Lin and Su(2021)]%
        {lin2021fast}
\bibfield{author}{\bibinfo{person}{Yi-Chung Lin} {and} \bibinfo{person}{Keh-Yih
  Su}.} \bibinfo{year}{2021}\natexlab{}.
\newblock \showarticletitle{How Fast can BERT Learn Simple Natural Language
  Inference?}. In \bibinfo{booktitle}{\emph{Proceedings of the 16th Conference
  of the European Chapter of the Association for Computational Linguistics:
  Main Volume}}. \bibinfo{pages}{626--633}.
\newblock


\bibitem[Liu et~al\mbox{.}(2023)]%
        {liu2023pre}
\bibfield{author}{\bibinfo{person}{Pengfei Liu}, \bibinfo{person}{Weizhe Yuan},
  \bibinfo{person}{Jinlan Fu}, \bibinfo{person}{Zhengbao Jiang},
  \bibinfo{person}{Hiroaki Hayashi}, {and} \bibinfo{person}{Graham Neubig}.}
  \bibinfo{year}{2023}\natexlab{}.
\newblock \showarticletitle{Pre-train, prompt, and predict: A systematic survey
  of prompting methods in natural language processing}.
\newblock \bibinfo{journal}{\emph{Comput. Surveys}} \bibinfo{volume}{55},
  \bibinfo{number}{9} (\bibinfo{year}{2023}), \bibinfo{pages}{1--35}.
\newblock


\bibitem[Liu et~al\mbox{.}(2019)]%
        {liu2019RoBERTa}
\bibfield{author}{\bibinfo{person}{Yinhan Liu}, \bibinfo{person}{Myle Ott},
  \bibinfo{person}{Naman Goyal}, \bibinfo{person}{Jingfei Du},
  \bibinfo{person}{Mandar Joshi}, \bibinfo{person}{Danqi Chen},
  \bibinfo{person}{Omer Levy}, \bibinfo{person}{Mike Lewis},
  \bibinfo{person}{Luke Zettlemoyer}, {and} \bibinfo{person}{Veselin
  Stoyanov}.} \bibinfo{year}{2019}\natexlab{}.
\newblock \showarticletitle{Roberta: A robustly optimized bert pretraining
  approach}.
\newblock \bibinfo{journal}{\emph{arXiv preprint arXiv:1907.11692}}
  (\bibinfo{year}{2019}).
\newblock


\bibitem[Mai et~al\mbox{.}(2022)]%
        {mai2022hybrid}
\bibfield{author}{\bibinfo{person}{Sijie Mai}, \bibinfo{person}{Ying Zeng},
  \bibinfo{person}{Shuangjia Zheng}, {and} \bibinfo{person}{Haifeng Hu}.}
  \bibinfo{year}{2022}\natexlab{}.
\newblock \showarticletitle{Hybrid contrastive learning of tri-modal
  representation for multimodal sentiment analysis}.
\newblock \bibinfo{journal}{\emph{IEEE Transactions on Affective Computing}}
  (\bibinfo{year}{2022}).
\newblock


\bibitem[Miranda-Correa et~al\mbox{.}(2018)]%
        {miranda2018amigos}
\bibfield{author}{\bibinfo{person}{Juan~Abdon Miranda-Correa},
  \bibinfo{person}{Mojtaba~Khomami Abadi}, \bibinfo{person}{Nicu Sebe}, {and}
  \bibinfo{person}{Ioannis Patras}.} \bibinfo{year}{2018}\natexlab{}.
\newblock \showarticletitle{Amigos: A dataset for affect, personality and mood
  research on individuals and groups}.
\newblock \bibinfo{journal}{\emph{IEEE Transactions on Affective Computing}}
  \bibinfo{volume}{12}, \bibinfo{number}{2} (\bibinfo{year}{2018}),
  \bibinfo{pages}{479--493}.
\newblock


\bibitem[Okada et~al\mbox{.}(2016)]%
        {okada2016estimating}
\bibfield{author}{\bibinfo{person}{Shogo Okada}, \bibinfo{person}{Yoshihiko
  Ohtake}, \bibinfo{person}{Yukiko~I Nakano}, \bibinfo{person}{Yuki Hayashi},
  \bibinfo{person}{Hung-Hsuan Huang}, \bibinfo{person}{Yutaka Takase}, {and}
  \bibinfo{person}{Katsumi Nitta}.} \bibinfo{year}{2016}\natexlab{}.
\newblock \showarticletitle{Estimating communication skills using dialogue acts
  and nonverbal features in multiple discussion datasets}. In
  \bibinfo{booktitle}{\emph{Proceedings of the 18th ACM International
  Conference on Multimodal Interaction}}. \bibinfo{pages}{169--176}.
\newblock


\bibitem[Pampouchidou et~al\mbox{.}(2017)]%
        {pampouchidou2017automatic}
\bibfield{author}{\bibinfo{person}{Anastasia Pampouchidou},
  \bibinfo{person}{Panagiotis~G Simos}, \bibinfo{person}{Kostas Marias},
  \bibinfo{person}{Fabrice Meriaudeau}, \bibinfo{person}{Fan Yang},
  \bibinfo{person}{Matthew Pediaditis}, {and} \bibinfo{person}{Manolis
  Tsiknakis}.} \bibinfo{year}{2017}\natexlab{}.
\newblock \showarticletitle{Automatic assessment of depression based on visual
  cues: A systematic review}.
\newblock \bibinfo{journal}{\emph{IEEE Transactions on Affective Computing}}
  \bibinfo{volume}{10}, \bibinfo{number}{4} (\bibinfo{year}{2017}),
  \bibinfo{pages}{445--470}.
\newblock


\bibitem[Poria et~al\mbox{.}(2019)]%
        {poria2019meld}
\bibfield{author}{\bibinfo{person}{Soujanya Poria}, \bibinfo{person}{Devamanyu
  Hazarika}, \bibinfo{person}{Navonil Majumder}, \bibinfo{person}{Gautam Naik},
  \bibinfo{person}{Erik Cambria}, {and} \bibinfo{person}{Rada Mihalcea}.}
  \bibinfo{year}{2019}\natexlab{}.
\newblock \showarticletitle{MELD: A Multimodal Multi-Party Dataset for Emotion
  Recognition in Conversations}. In \bibinfo{booktitle}{\emph{Proceedings of
  the 57th Annual Meeting of the Association for Computational Linguistics}}.
  \bibinfo{pages}{527--536}.
\newblock


\bibitem[Schuller et~al\mbox{.}(2009)]%
        {schuller2009interspeech}
\bibfield{author}{\bibinfo{person}{B Schuller}, \bibinfo{person}{S Steidl},
  {and} \bibinfo{person}{A Batliner}.} \bibinfo{year}{2009}\natexlab{}.
\newblock \showarticletitle{The Interspeech 2009 Emotion Challenge}. In
  \bibinfo{booktitle}{\emph{Proc. Interspeech 2009, Brighton, UK}}.
  \bibinfo{pages}{312--315}.
\newblock


\bibitem[Soleymani et~al\mbox{.}(2017)]%
        {soleymani2017survey}
\bibfield{author}{\bibinfo{person}{Mohammad Soleymani}, \bibinfo{person}{David
  Garcia}, \bibinfo{person}{Brendan Jou}, \bibinfo{person}{Bj{\"o}rn Schuller},
  \bibinfo{person}{Shih-Fu Chang}, {and} \bibinfo{person}{Maja Pantic}.}
  \bibinfo{year}{2017}\natexlab{}.
\newblock \showarticletitle{A survey of multimodal sentiment analysis}.
\newblock \bibinfo{journal}{\emph{Image and Vision Computing}}
  \bibinfo{volume}{65} (\bibinfo{year}{2017}), \bibinfo{pages}{3--14}.
\newblock


\bibitem[Wang et~al\mbox{.}(2021a)]%
        {wang2021entailment}
\bibfield{author}{\bibinfo{person}{Sinong Wang}, \bibinfo{person}{Han Fang},
  \bibinfo{person}{Madian Khabsa}, \bibinfo{person}{Hanzi Mao}, {and}
  \bibinfo{person}{Hao Ma}.} \bibinfo{year}{2021}\natexlab{a}.
\newblock \showarticletitle{Entailment as few-shot learner}.
\newblock \bibinfo{journal}{\emph{arXiv preprint arXiv:2104.14690}}
  (\bibinfo{year}{2021}).
\newblock


\bibitem[Wang et~al\mbox{.}(2019)]%
        {wang2019structBERT}
\bibfield{author}{\bibinfo{person}{Wei Wang}, \bibinfo{person}{Bin Bi},
  \bibinfo{person}{Ming Yan}, \bibinfo{person}{Chen Wu}, \bibinfo{person}{Zuyi
  Bao}, \bibinfo{person}{Jiangnan Xia}, \bibinfo{person}{Liwei Peng}, {and}
  \bibinfo{person}{Luo Si}.} \bibinfo{year}{2019}\natexlab{}.
\newblock \showarticletitle{Structbert: Incorporating language structures into
  pre-training for deep language understanding}.
\newblock \bibinfo{journal}{\emph{arXiv preprint arXiv:1908.04577}}
  (\bibinfo{year}{2019}).
\newblock


\bibitem[Wang et~al\mbox{.}(2021b)]%
        {wang2021m2lens}
\bibfield{author}{\bibinfo{person}{Xingbo Wang}, \bibinfo{person}{Jianben He},
  \bibinfo{person}{Zhihua Jin}, \bibinfo{person}{Muqiao Yang},
  \bibinfo{person}{Yong Wang}, {and} \bibinfo{person}{Huamin Qu}.}
  \bibinfo{year}{2021}\natexlab{b}.
\newblock \showarticletitle{M2Lens: visualizing and explaining multimodal
  models for sentiment analysis}.
\newblock \bibinfo{journal}{\emph{IEEE Transactions on Visualization and
  Computer Graphics}} \bibinfo{volume}{28}, \bibinfo{number}{1}
  (\bibinfo{year}{2021}), \bibinfo{pages}{802--812}.
\newblock


\bibitem[Wei et~al\mbox{.}(2022)]%
        {wei2022investigating}
\bibfield{author}{\bibinfo{person}{Wenqing Wei}, \bibinfo{person}{Sixia Li},
  {and} \bibinfo{person}{Shogo Okada}.} \bibinfo{year}{2022}\natexlab{}.
\newblock \showarticletitle{Investigating the relationship between dialogue and
  exchange-level impression}. In \bibinfo{booktitle}{\emph{Proceedings of the
  2022 International Conference on Multimodal Interaction}}.
  \bibinfo{pages}{359--367}.
\newblock


\bibitem[Wei et~al\mbox{.}(2021)]%
        {wei2021multimodal}
\bibfield{author}{\bibinfo{person}{Wenqing Wei}, \bibinfo{person}{Sixia Li},
  \bibinfo{person}{Shogo Okada}, {and} \bibinfo{person}{Kazunori Komatani}.}
  \bibinfo{year}{2021}\natexlab{}.
\newblock \showarticletitle{Multimodal user satisfaction recognition for
  non-task oriented dialogue systems}. In \bibinfo{booktitle}{\emph{Proceedings
  of the 2021 International Conference on Multimodal Interaction}}.
  \bibinfo{pages}{586--594}.
\newblock


\bibitem[Yang et~al\mbox{.}(2020)]%
        {yang2020cm}
\bibfield{author}{\bibinfo{person}{Kaicheng Yang}, \bibinfo{person}{Hua Xu},
  {and} \bibinfo{person}{Kai Gao}.} \bibinfo{year}{2020}\natexlab{}.
\newblock \showarticletitle{Cm-bert: Cross-modal bert for text-audio sentiment
  analysis}. In \bibinfo{booktitle}{\emph{Proceedings of the 28th ACM
  international conference on multimedia}}. \bibinfo{pages}{521--528}.
\newblock


\bibitem[Yang et~al\mbox{.}(2018)]%
        {yang2018commonsense}
\bibfield{author}{\bibinfo{person}{Shaohua Yang}, \bibinfo{person}{Qiaozi Gao},
  \bibinfo{person}{Sari Saba-Sadiya}, {and} \bibinfo{person}{Joyce Chai}.}
  \bibinfo{year}{2018}\natexlab{}.
\newblock \showarticletitle{Commonsense justification for action explanation}.
  In \bibinfo{booktitle}{\emph{Proceedings of the 2018 Conference on Empirical
  Methods in Natural Language Processing}}. \bibinfo{pages}{2627--2637}.
\newblock


\bibitem[Yang et~al\mbox{.}(2022)]%
        {yang2022few}
\bibfield{author}{\bibinfo{person}{Xiaocui Yang}, \bibinfo{person}{Shi Feng},
  \bibinfo{person}{Daling Wang}, \bibinfo{person}{Pengfei Hong}, {and}
  \bibinfo{person}{Soujanya Poria}.} \bibinfo{year}{2022}\natexlab{}.
\newblock \showarticletitle{Few-shot Multimodal Sentiment Analysis based on
  Multimodal Probabilistic Fusion Prompts}.
\newblock \bibinfo{journal}{\emph{arXiv preprint arXiv:2211.06607}}
  (\bibinfo{year}{2022}).
\newblock


\bibitem[Yu et~al\mbox{.}(2020)]%
        {yu2020ch}
\bibfield{author}{\bibinfo{person}{Wenmeng Yu}, \bibinfo{person}{Hua Xu},
  \bibinfo{person}{Fanyang Meng}, \bibinfo{person}{Yilin Zhu},
  \bibinfo{person}{Yixiao Ma}, \bibinfo{person}{Jiele Wu},
  \bibinfo{person}{Jiyun Zou}, {and} \bibinfo{person}{Kaicheng Yang}.}
  \bibinfo{year}{2020}\natexlab{}.
\newblock \showarticletitle{Ch-sims: A chinese multimodal sentiment analysis
  dataset with fine-grained annotation of modality}. In
  \bibinfo{booktitle}{\emph{Proceedings of the 58th annual meeting of the
  association for computational linguistics}}. \bibinfo{pages}{3718--3727}.
\newblock


\bibitem[Yuan et~al\mbox{.}(2021)]%
        {yuan2021survey}
\bibfield{author}{\bibinfo{person}{Jun Yuan}, \bibinfo{person}{Changjian Chen},
  \bibinfo{person}{Weikai Yang}, \bibinfo{person}{Mengchen Liu},
  \bibinfo{person}{Jiazhi Xia}, {and} \bibinfo{person}{Shixia Liu}.}
  \bibinfo{year}{2021}\natexlab{}.
\newblock \showarticletitle{A survey of visual analytics techniques for machine
  learning}.
\newblock \bibinfo{journal}{\emph{Computational Visual Media}}
  \bibinfo{volume}{7} (\bibinfo{year}{2021}), \bibinfo{pages}{3--36}.
\newblock


\bibitem[Zadeh et~al\mbox{.}(2017)]%
        {zadeh2017tensor}
\bibfield{author}{\bibinfo{person}{Amir Zadeh}, \bibinfo{person}{Minghai Chen},
  \bibinfo{person}{Soujanya Poria}, \bibinfo{person}{Erik Cambria}, {and}
  \bibinfo{person}{Louis-Philippe Morency}.} \bibinfo{year}{2017}\natexlab{}.
\newblock \showarticletitle{Tensor Fusion Network for Multimodal Sentiment
  Analysis}. In \bibinfo{booktitle}{\emph{Proceedings of the 2017 Conference on
  Empirical Methods in Natural Language Processing}}.
  \bibinfo{pages}{1103--1114}.
\newblock


\bibitem[Zadeh et~al\mbox{.}(2016)]%
        {zadeh2016mosi}
\bibfield{author}{\bibinfo{person}{Amir Zadeh}, \bibinfo{person}{Rowan
  Zellers}, \bibinfo{person}{Eli Pincus}, {and} \bibinfo{person}{Louis-Philippe
  Morency}.} \bibinfo{year}{2016}\natexlab{}.
\newblock \showarticletitle{Mosi: multimodal corpus of sentiment intensity and
  subjectivity analysis in online opinion videos}.
\newblock \bibinfo{journal}{\emph{arXiv preprint arXiv:1606.06259}}
  (\bibinfo{year}{2016}).
\newblock


\bibitem[Zou et~al\mbox{.}(2022)]%
        {zou2022utilizing}
\bibfield{author}{\bibinfo{person}{Wenwen Zou}, \bibinfo{person}{Jundi Ding},
  {and} \bibinfo{person}{Chao Wang}.} \bibinfo{year}{2022}\natexlab{}.
\newblock \showarticletitle{Utilizing BERT Intermediate Layers for Multimodal
  Sentiment Analysis}. In \bibinfo{booktitle}{\emph{2022 IEEE International
  Conference on Multimedia and Expo (ICME)}}. IEEE, \bibinfo{pages}{1--6}.
\newblock


\end{thebibliography}

\end{document}